\documentclass[10pt,twocolumn,letterpaper]{article}
\usepackage{xcolor}

\usepackage[pagenumbers]{cvpr} 
\usepackage{tabularx}
\usepackage{multirow}

\definecolor{cvprblue}{rgb}{0.21,0.49,0.74}
\usepackage[pagebackref,breaklinks,colorlinks,allcolors=cvprblue]{hyperref}

\def\paperID{8376} 
\def\confName{CVPR}
\def\confYear{2026}

\title{DReX: Pure Vision Fusion of Self-Supervised and Convolutional Representations for Image Complexity Prediction}

\author{
Jonathan Skaza$^\dagger$  \quad Parsa Madinei \quad Ziqi Wen \quad Miguel Eckstein\\
University of California, Santa Barbara \\
$^\dagger$ Corresponding author: \normalsize \texttt{skaza@ucsb.edu}
}

\begin{document}
\maketitle
\begin{abstract}
  Visual complexity prediction is a fundamental problem in computer vision with applications in image compression, retrieval, and classification. Understanding what makes humans perceive an image as complex is also a long-standing question in cognitive science. Recent approaches have leveraged multimodal models that combine visual and linguistic representations, but it remains unclear whether language information is necessary for this task. We propose DReX (DINO-ResNet Fusion), a vision-only model that fuses self-supervised and convolutional representations through a learnable attention mechanism to predict image complexity. Our architecture integrates multi-scale hierarchical features from ResNet-50 with semantically rich representations from DINOv3 ViT-S/16, enabling the model to capture both low-level texture patterns and high-level semantic structure. DReX achieves state-of-the-art performance on the IC9600 benchmark (Pearson $r = 0.9581$), surpassing previous methods—including those trained on multimodal image-text data—while using approximately $21.5\times$ fewer learnable parameters. Furthermore, DReX generalizes robustly across multiple datasets and metrics, achieving superior results on Pearson and Spearman correlation, Root Mean Square Error (RMSE), and Mean Absolute Error (MAE). Ablation and attention analyses confirm that DReX leverages complementary cues from both backbones, with the DINOv3 [CLS] token enhancing sensitivity to visual complexity. Our findings suggest that visual features alone can be sufficient for human-aligned complexity prediction and that, when properly fused, self-supervised transformers and supervised deep convolutional neural networks offer complementary and synergistic benefits for this task.
  \end{abstract}

\begin{figure*}[t]
  \centering
  \includegraphics[width=\linewidth]{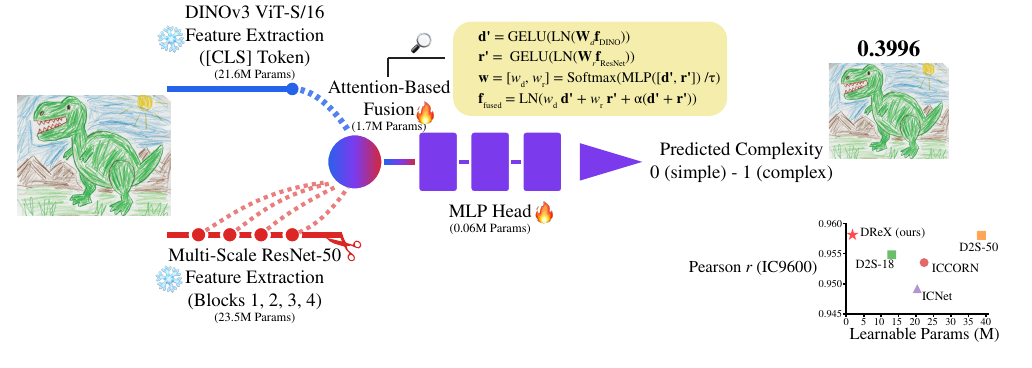}
  \caption{\textbf{DReX architecture for visual complexity prediction.} Multi-scale ResNet features and DINOv3's [CLS] embedding are adaptively fused via learned attention. ResNet blocks 1–4 provide hierarchical spatial information at different scales, while DINOv3’s [CLS] token captures global semantic content. The attention fusion module learns to weight each representation, and an MLP head regresses the complexity score. Snowflakes indicate frozen pretrained weights; flames indicate trainable components ($\sim$1.8M parameters).}
  \label{fig:1}
  \end{figure*}
\section{Introduction}
\label{sec:intro}

What makes one image more complex than another? This deceptively simple question touches on fundamental issues in perception, cognition, and computation. A child's crayon drawing feels simpler than an Escher illustration filled with intricate patterns and impossible geometries, yet predicting this intuitive judgment computationally has challenged researchers for decades~\cite{birkhoff1933aesthetic,chipman1977complexity,forsythe2009visual}. Visual complexity reflects not only low-level properties such as edge density or color variation, but also the interplay between structural regularities, statistical expectations, and semantic content~\cite{kyle2023characterising,oliva2004identifying}. Rooted in Gestalt psychology’s principle that perception favors simple ``good'' forms (prägnanz)~\cite{koffka1935principles,donderi2006visual,van2024pragnanz}, and linked to information theory's notion that simpler patterns admit more concise descriptions~\cite{heaps1999similarity}, this relationship between perceptual simplicity and descriptive efficiency provides a foundation for computational measures of visual complexity.

Early computer vision approaches quantified complexity using low-level statistics---including edge density~\cite{rosenholtz2007measuring,dai2022visual}, number of regions~\cite{comaniciu2002mean}, color variety~\cite{guo2018assessment}, feature congestion~\cite{rosenholtz2007measuring}, and subband entropy~\cite{rosenholtz2007measuring}---or algorithmic measures such as file compression ratios~\cite{donderi2006visual,machado2015computerized}. Yet, human judgments emerge from multi-scale interactions: local textures, symmetry, and repetition combine with global organization and semantic factors such as object count and scene clutter~\cite{oliva2004identifying}. 

Modern deep neural networks (DNNs) inherently capture hierarchical visual representations. Deep convolutional neural networks (DCNNs), for instance, integrate low-level statistics with higher-level abstractions, enabling data-driven predictions of perceived complexity~\cite{saraee2020visual}. Specialized architectures such as ICNet~\cite{feng2022ic9600} and Describe-to-Score (D2S)~\cite{liu2025describe} have achieved strong performance on large-scale visual complexity benchmarks such as IC9600~\cite{feng2022ic9600}, demonstrating that deep hierarchical features can approximate human complexity judgments.

A key question remains: are multimodal or end-to-end task-specific models truly necessary, or can vision-only representations suffice for predicting human complexity? Many recent approaches rely on language supervision or extensive task-specific training. In contrast, self-supervised vision transformers such as DINOv3 learn semantically organized features from images alone~\cite{simeoni2025dinov3}, indicating that pretrained visual representations may already encode the structure relevant to human perception. In this work, we test the hypothesis that vision-only pretrained DNNs, with minimal task-specific adaptation, can accurately predict human judgments of visual complexity.

We compare several DNN models on the IC9600 benchmark to evaluate this hypothesis. Our analysis shows that merging self-supervised DINO embeddings with hierarchical DCNN features achieves state-of-the-art performance among deep learning methods, even without multimodal pretraining or end-to-end task-specific training. Building on this insight, we introduce DReX (DINO-ResNet Fusion), a vision-only architecture that fuses multi-scale ResNet-50 features~\cite{he2016deep} with DINOv3's global [CLS] embedding via a learnable attention mechanism. This design integrates fine-grained spatial detail with global semantic context, achieving top performance on the IC9600 benchmark while remaining entirely vision-based—without textual, multimodal, or handcrafted inputs.

These findings highlight the importance of representation quality in modeling visual complexity. DReX leverages DINO's emergent semantic organization together with pretrained hierarchical features from ResNet-50, using both as fixed backbones rather than training them directly. By building on these representations, DReX delivers a highly effective and extensible vision-only framework for predicting visual complexity.

The contributions of this paper can be summarized as follows:
\begin{itemize}
	\item We introduce DReX, a novel fusion of two pretrained visual backbones that achieves leading performance in human-aligned visual complexity prediction with $\sim$21.5$\times$ fewer learnable parameters than the most comparably performing model.
	\item Unlike prior models that rely on language supervision or heavy finetuning, DReX reads out features from two non-finetuned visual backbones, keeping the trained portion lightweight.
	\item We analyze the utility of the DINOv3 [CLS] token in visual complexity prediction, showing that while it is not sufficient on its own for visual complexity prediction, it provides complementary value as a fusion component.
\end{itemize}

\section{Related Work}
\label{sec:related-work}
\subsection{Datasets for Visual Complexity Assessment}
The development of robust models for visual complexity prediction relies heavily on high-quality, annotated datasets that capture human judgments across diverse image types. Early efforts in this area were limited by small-scale datasets, but recent benchmarks have expanded in size and variety to better reflect real-world perceptual challenges.

The SAVOIAS dataset~\cite{saraee2018savoias} represents one of the first multi-category collections specifically designed for visual complexity research. It comprises 1,420 images spanning seven categories, including scenes, advertisements, visualizations/infographics, objects, interior design, art, and suprematism. Images were sourced from existing datasets and annotated by human raters for complexity levels, providing a diverse foundation for evaluating complexity metrics.

Building on this, the IC9600 dataset~\cite{feng2022ic9600} introduced a considerably larger benchmark with 9,600 annotated images drawn from eight categories: abstract art, advertisements, architecture, objects, paintings, persons, scenes, and transport. Annotated by 17 human raters, IC9600 emphasizes comprehensive coverage of complexity variations and has become a standard for training and evaluating deep learning models.

Additionally, adaptations of established computer vision datasets have been employed for visual complexity studies. The PASCAL VOC dataset~\cite{everingham2010pascal}, originally used for object detection, has been repurposed with complexity annotations in subsets like Nagle4k~\cite{nagle2020predicting}, which includes 4,000 images rated for visual complexity.

These datasets collectively address key challenges in visual complexity research by providing varied, human-annotated resources that support the training of models capable of aligning with perceptual judgments.

\subsection{Modern Deep Learning Approaches}

Recent advances in visual complexity prediction have moved beyond hand-crafted features, embracing deep learning architectures that automatically extract hierarchical, multi-scale representations. These methods consistently outperform traditional low-level metrics.

ICNet~\cite{feng2022ic9600}, introduced alongside the IC9600 dataset, leverages convolutional layers to capture both local details and global structure, achieving strong baseline performance on large-scale benchmarks and demonstrating the effectiveness of vision-only, end-to-end learning for the task. Building on this, ICCORN~\cite{guo2023image} integrates ICNet with deep ordinal regression through the Conditional Ordinal Regression Network (CORN), framing complexity prediction as an ordinal ranking problem and improving evaluation accuracy.

In the realm of unsupervised learning, CLICv2~\cite{liu2025clicv2} advances contrastive learning paradigms for image complexity representation. By enforcing content invariance through contrastive objectives, it learns meaningful complexity features from unlabeled data.

Multimodal approaches have further enriched the field. Describe-to-Score (D2S)~\cite{liu2025describe} leverages pretrained vision-language models to generate descriptive captions and derive complexity scores partially from textual representations. This text-guided approach achieves strong results on IC9600 and demonstrates robust cross-dataset generalization to SAVOIAS and PASCAL VOC.

Overall, recent years have seen progressive performance gains across datasets and model families, but these improvements often come hand-in-hand with increasingly intricate architectures and elaborate training regimes. This trend highlights a growing trade-off between evaluation performance and pipeline complexity, motivating the need for simpler, vision-only approaches that achieve competitive performance without relying on heavy multimodal pipelines or specialized optimization strategies. 

\section{Methods}
\label{sec:methods}
We propose DReX, a fusion architecture that combines self-supervised DINOv3 features with pretrained multi-scale ResNet representations to predict image complexity as perceived by human raters. Formally, the task is to learn a function $f_\theta: \mathcal{X} \rightarrow \mathbb{R} $ that maps an input image $x \in \mathcal{X}$ to its predicted human-rated complexity $\hat{c} = f_\theta(x)$. The model leverages two frozen backbones (DINOv3 ViT-S/16 and ResNet-50) for feature extraction, followed by an attention-based fusion module and a lightweight regression head (Figure ~\ref{fig:1}).

\subsection{Feature Extraction Backbones}
\subsubsection{DINO Backbone} 
We employ the pretrained DINOv3 ViT-S/16 model~\cite{simeoni2025dinov3} as an encoder to extract high-level semantic features from images. DINOv3 is a state-of-the-art self-supervised learning framework that trains Vision Transformers (ViT) models on large, curated datasets without labels, using global and local objectives to produce robust representations. The ViT-S/16 variant, the smallest model ($\sim$21.6M parameters) in the DINOv3 family, is a distilled version of the full DINOv3 model pretrained on the LVD-1689M dataset. In our experiments, we observed no performance gains when using larger models or higher-dimensional embeddings (see Supplementary 1). 

For feature extraction, images are fed into the frozen backbone to extract the [CLS] token embedding from the last hidden state, yielding a feature vector, $\mathbf{f}_{\text{DINO}} \in \mathbb{R}^{384}$. This token is thought to compress global image context into vector space.

\subsubsection{Multi-Scale ResNet Backbone} 
To capture multi-resolution spatial details, we leverage ResNet-50~\cite{he2016deep}, a DCNN with residual blocks pretrained in a supervised manner on ImageNet~\cite{deng2009imagenet}.

Features are extracted after each residual stage (layers 1-4), globally average-pooled to $1 \times 1$, and concatenated, resulting in $\mathbf{f}_{\text{ResNet}} \in \mathbb{R}^{3840}$ (256+512+1024+2048 channels). This multi-scale approach captures hierarchical features from low-level edges to higher-level aggregations. Input images are resized to $512 \times 512$ and normalized per standard ResNet preprocessing.

\subsection{Attention-Based Fusion}
The DINO ($\mathbf{f}_{\text{DINO}}$) and ResNet ($\mathbf{f}_{\text{ResNet}}$) features are projected to independent hidden dimensions  via linear layers followed by LayerNorm and GELU activation. These projections, $\mathbf{d}^\prime$ and $\mathbf{r}^\prime$, are concatenated and passed through a two-layer MLP to compute unnormalized attention logits. Softmax with learnable temperature, $\tau$, yields weights $\mathbf{w} \in \mathbb{R}^2$:

\begin{equation}
\mathbf{w} = [w_d,w_r] = \text{Softmax}\left(\frac{\text{MLP}([\mathbf{d}^\prime, \mathbf{r}^\prime])}{\tau}\right)
\end{equation}

The fused representation is computed as:
\begin{equation}
  \label{eq:fusion}
\mathbf{f} = w_d \mathbf{d}^\prime + w_r \mathbf{r}^\prime + \alpha (\mathbf{d}^\prime + \mathbf{r}^\prime)
\end{equation}
where $\alpha$ is a learnable residual scaling. Final normalization applies LayerNorm to $\mathbf{f}$.
This mechanism adaptively weighs semantic (DINO) and structural (ResNet) cues, bringing both local and global context into a single representation.
\subsection{Regression Head and Inference}
A four-layer MLP head maps the fused features to a scalar complexity score. The architecture follows a progressive dimensionality reduction: $f \in \mathbb{R}^{384} \rightarrow \mathbb{R}^{128} \rightarrow \mathbb{R}^{64} \rightarrow \mathbb{R}^{32}\rightarrow \mathbb{R}^{1}$, with GELU activations after each hidden layer. This lightweight head ensures DReX relies primarily on pretrained feature quality rather than extensive task-specific adaptation.

\subsection{Training Details}
We trained DReX on the IC9600 training split, which consists of 6,720 images annotated for perceptual complexity across eight semantic categories. Only the fusion and head modules were updated during training, while all backbone components remained frozen. We adopted the Huber loss with an AdamW optimizer~\cite{loshchilov2017decoupled} (initial learning rate 0.001) and a OneCycleLR schedule over just 10 epochs with batch size of 16. An Exponential Moving Average (EMA, decay 0.999) was applied to stabilize the fusion and head parameters. We applied random dropout ($p=0.1$) between layers of the MLP head.

Compared to prior methods that rely on elaborate multi-part objectives, our training regime is intentionally simple. For example, ICNet employs a two-component loss to jointly optimize global and local complexity predictions, while D2S incorporates three losses—MSE regression, entropy alignment, and contrastive feature alignment—to enforce semantic consistency and cross-modal correspondence. In contrast, DReX achieves strong performance with a very simple training regime.

\subsection{Evaluation Metrics}

We evaluated the trained DReX on the IC9600 test split, which consists of an additional 2,880 images annotated for perceptual complexity across the eight semantic categories. To assess generalization capabilities, we also evaluated DReX on the SAVOIAS and PASCAL VOC (Nagle4k) datasets.

We report Pearson correlation coefficient ($r$), Spearman's rank correlation coefficient ($\rho$), Root Mean Squared Error (RMSE), and Mean Absolute Error (MAE) between the predicted and human-annotated complexity scores:

\begin{equation}
\text{Pearson } r = \frac{\sum_{i=1}^n (c_i - \bar{c})(\hat{c}_i - \bar{\hat{c}})}{\sqrt{\sum_{i=1}^n (c_i - \bar{c})^2 \sum_{i=1}^n (\hat{c}_i - \bar{\hat{c}})^2}}
\end{equation}
\begin{equation}
\text{RMSE} = \sqrt{\frac{1}{n} \sum_{i=1}^n (c_i - \hat{c}_i)^2}
\end{equation}
\begin{equation}
\text{MAE} = \frac{1}{n} \sum_{i=1}^n |c_i - \hat{c}_i|
\end{equation}

\noindent where $n$ is the number of images in the evaluation set, $c_i$ is the human-annotated complexity score, $\hat{c}_i$ is the model-predicted complexity score, and $\bar{c}$ and $\bar{\hat{c}}$ denote the mean human-annotated and mean model-predicted complexity scores, respectively. Spearman’s rank correlation coefficient, $\rho$, is computed by first converting the complexity scores to ranks and then applying the Pearson correlation formula to the ranked values.

\section{Results}
\label{sec:results}

\begin{table*}
  \caption{\textbf{Comparison of deep learning models for complexity prediction on the IC9600 dataset.} DReX outperforms previous methods, including those trained on multimodal image-text data, in terms of Pearson correlation between human and model-predicted complexity scores. While it slightly trails the D2S-R50 model in Spearman correlation, it outperforms the next best vision-only model, ICCORN.$^\dagger$}
  \label{tab:complexity-prediction-models}
  \centering
  \begin{tabularx}{\textwidth}{XXccc}
    \toprule
    \textbf{Method} & \textbf{Training Modality} & \textbf{Pearson $r$} $\uparrow$ & \textbf{Spearman's $\rho$} $\uparrow$ & \textbf{Params} $\downarrow$ \\
    \midrule
    DReX (ours) & Vision & \textbf{0.9581} & 0.9542 & \textbf{1.8M}\\
    D2S-R50~\cite{liu2025describe} & Vision + Text & 0.9580 & \textbf{0.9544} & 38.7M\\
    D2S-R18~\cite{liu2025describe} & Vision + Text & 0.9548  & 0.9521  & 13M\\
    ICCORN$^\ddagger$~\cite{guo2023image} & Vision &  0.9535 & 0.9510  & 22.3M \\
    ICNet~\cite{feng2022ic9600} & Vision &  0.9492 &  0.9450 & 20.3M\\
    CLICv2$^\ddagger$~\cite{liu2025clicv2} & Text & 0.9330 & 0.9270 &  \_ \\
    \bottomrule
  \end{tabularx}
  \textsuperscript{$^\dagger$Pearson $r$ and Spearman’s $\rho$ are designated as the primary metrics because they are the measures common to all studies.}

  \textsuperscript{$^\ddagger$ indicates that we were unable to reproduce the implementation; in these cases, we include the reported results from the original paper.}
\end{table*}

\begin{table*}[t]
  \caption{\textbf{Comparison of deep learning models for complexity prediction across datasets.} DReX achieves the best performance on the standard evaluation metrics for the external datasets SAVOIAS and PASCAL VOC, demonstrating its strong generalization beyond IC9600. On IC9600, DReX remains highly competitive across all metrics.}  \label{tab:performance-comparison}
  \centering
  \begin{tabularx}{\textwidth}{l l *{4}{>{\centering\arraybackslash}X}}
    \toprule
    \textbf{Dataset} & \textbf{Method} & \textbf{Pearson $r$} $\uparrow$ & \textbf{Spearman's $\rho$} $\uparrow$ & \textbf{RMSE} $\downarrow$ & \textbf{RMAE} $\downarrow$ \\
    \midrule
    \multirow{3}{*}{IC9600~\cite{feng2022ic9600}} 
      & DReX (ours) & \textbf{0.9581} & 0.9542 & 0.0520 & 0.2031 \\
      & D2S-R50~\cite{liu2025describe} & 0.9580 & \textbf{0.9544} & 0.0497 & 0.1966 \\
      & D2S-R18~\cite{liu2025describe} & 0.9548 & 0.9521 & \textbf{0.0496} & \textbf{0.1963} \\
      & ICNet~\cite{feng2022ic9600} & 0.9492 & 0.9450 & 0.0530 & 0.2037 \\
    \midrule
    \multirow{3}{*}{SAVOIAS$^\dagger$ \cite{saraee2018savoias}} 
      & DReX (ours) &  \textbf{0.8574} &  \textbf{0.8596} &  \textbf{0.1775} &  \textbf{0.3798} \\
      & D2S-R50~\cite{liu2025describe} &  0.8569 &  0.8570 & 0.1798 &  0.3823 \\
      & D2S-R18~\cite{liu2025describe} &  0.8413 &  0.8414 & 0.1805 & 0.3834 \\
      & ICNet~\cite{feng2022ic9600} &  0.8492 & 0.8520 &  0.1820 &  0.3851 \\
    \midrule
    \multirow{3}{*}{PASCAL VOC$^\ddagger$ \cite{nagle2020predicting}} 
      & DReX (ours) & \textbf{0.7839} & \textbf{0.8038}  &  \_& \_ \\
      & D2S-R50~\cite{liu2025describe} & 0.7765  & 0.7976 & \_ & \_ \\
      & D2S-R18~\cite{liu2025describe} &  0.7748 & 0.7976 & \_ & \_ \\
      & ICNet~\cite{feng2022ic9600} &  0.7666 &    0.7851 & \_ & \_ \\
    \bottomrule
  \end{tabularx}
  \textsuperscript{$^\dagger$ Reported metric values are the mean of the metric across the categories in SAVOIAS, consistent with \cite{feng2022ic9600}.}

  \textsuperscript{$^\ddagger$ RMSE and RMAE are not reported for PASCAL VOC, as the dataset contains a different range of complexity scores (not 0-1 as in IC9600 and SAVOIAS).}
\end{table*}

\subsection{Performance on IC9600}
We evaluate DReX on the IC9600 test set and compare against other top deep learning methods (Table~\ref{tab:complexity-prediction-models}). DReX achieves a Pearson correlation of $r = 0.9581$, exceeding prior methods, including multimodal approaches. Notably, DReX marginally outperforms D2S-R50 ($r = 0.9580$), which leverages vision-language pretraining, while using $\sim$21.5$\times$ fewer trainable parameters ($\sim$1.8M vs $\sim$38.7M) (see Supplementary 2 for seed-to-seed performance stability.)

On Spearman's rank correlation, DReX achieves $\rho = 0.9542$, marginally below D2S-R50's $\rho = 0.9544$ but substantially outperforming the vision-only baselines. Compared to ICCORN, the strongest vision-only competitor ($r = 0.9535$, $\rho = 0.9510$, $\sim$22.3M params), DReX demonstrates consistent improvements across both correlation metrics.

These results establish that pure vision representations, when appropriately fused, can achieve leading performance on visual complexity prediction without requiring language supervision or extensive parameter tuning.

\subsection{Cross-Dataset Generalization}
To assess generalization beyond IC9600, we evaluate DReX on two other benchmarks: SAVOIAS and PASCAL VOC (Table~\ref{tab:performance-comparison}). On SAVOIAS, DReX achieves the strongest performance across all metrics and reproduced models, with $r = 0.8574$, $\rho = 0.8596$, RMSE = 0.1775, and RMAE = 0.3798. On PASCAL VOC, DReX similarly outperforms competitors with $r = 0.7839$ and $\rho = 0.8038$.

Notably, while DReX achieves comparable performance to D2S models on IC9600 (surpassing in $r$; though it achieves slightly lower performance on $\rho$ and residual error metrics) it exhibits stronger generalization to different datasets. This behavior aligns with DReX's architectural design; by combining frozen foundation models through a lightweight fusion mechanism with $\sim$1.8M parameters, the model learns to aggregate complementary pretrained representations without overfitting to training set-specific artifacts. In contrast, methods requiring end-to-end tuning of larger networks may achieve marginal in-distribution gains at the expense of reduced generalization. These results indicate that accessing pretrained self-supervised and supervised visual features through minimal trainable parameters provides a robust foundation for complexity prediction that generalizes more effectively than approaches relying on vision-language alignment or extensive task-specific adaptation.

\section{Experiments}
\label{sec:experiments}
\subsection{Branch Ablation Studies}
\begin{figure*}[t]
  \centering
   \includegraphics[width=\linewidth]{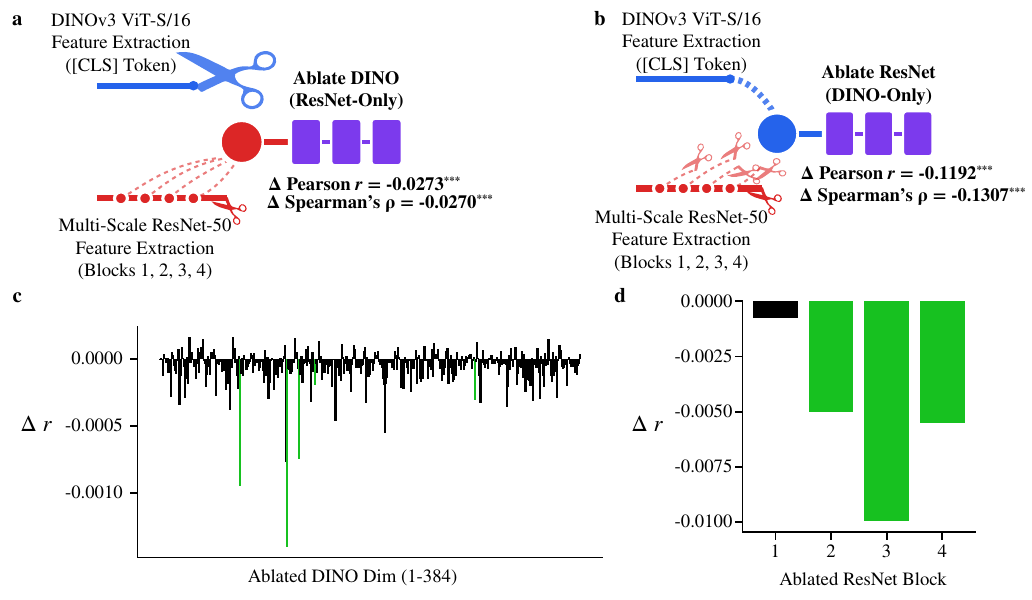}
   \caption{\textbf{Ablation experiments on the IC9600 test set.} 
   (a) Change in correlation when ablating the entire DINO branch from DReX.
   (b) Change in correlation when ablating the entire ResNet branch from DReX.
   (c) Change in Pearson correlation ($r$) when ablating individual DINO embedding dimensions. Green lines indicate FDR-corrected significance at $p<0.01$. 
   (d) Change in Pearson correlation ($r$) when ablating individual ResNet feature dimensions. Green bars indicate FDR-corrected significance at $p<0.01$. $^{***}$ denotes permutation test significance at $p<0.001$.}
  \label{fig:fig2}
\end{figure*}
Using the IC9600 test set, we conducted ablations to quantify the contribution of each pretrained backbone branch to DReX's performance (Figure~\ref{fig:fig2}a,b). To isolate each component's effect, we zero the features from the ablated branch while preserving the model's learned attention mechanism and residual connections. This approach tests how the model naturally responds when one feature extractor provides uninformative (zero) features:
\begin{equation}
\label{eq:ablation-dino}
\mathbf{f_{-DINO}} = w_d \mathbf{d}^\prime_{\text{zero}} + w_r \mathbf{r}^\prime + \alpha (\mathbf{d}^\prime_{\text{zero}} + \mathbf{r}^\prime)
\end{equation}
\begin{equation}
\label{eq:ablation-resnet}
\mathbf{f_{-ResNet}} = w_d \mathbf{d}^\prime + w_r \mathbf{r}^\prime_{\text{zero}} + \alpha (\mathbf{d}^\prime + \mathbf{r}^\prime_{\text{zero}})
\end{equation}
where $\mathbf{d}^\prime_{\text{zero}}$ and $\mathbf{r}^\prime_{\text{zero}}$ denote zero embeddings, and attention weights $w_d,w_r$ are computed by the model in response to the zeroed features.

Ablating the DINO branch (Equation~\ref{eq:ablation-dino}) yields Pearson $r=0.9307$ ($\Delta r = -0.0273$) and Spearman's $\rho = 0.9271$ ($\Delta \rho = -0.0270$)
 both slightly below the full DReX model. Conversely, ablating the ResNet branch (Equation~\ref{eq:ablation-resnet}) produces Pearson $r=0.8389$ ($\Delta r = -0.1192$) and Spearman's $\rho = 0.8235$ ($\Delta \rho = -0.1307$), reflecting a more substantial drop.

Although the magnitude of degradation differs between branches, both ablations result in statistically significant performance decreases (permutation test, $p < 0.001$). These findings suggest that both backbones contribute meaningfully to DReX's effectiveness and that the fusion mechanism successfully exploits their complementary strengths. The larger drop from ResNet ablation indicates it captures essential features that predict image complexity, while DINO's contribution, though smaller, provides complementary information.

While previous methods primarily relied on ResNet-based architectures, DReX shows that DINOv3's [CLS] token provides a useful additional signal, offering a small but non-negligible boost in predicting visual complexity.

\subsection{Feature Dimension Ablation Studies}
\begin{figure*}[t]
  \centering
   \includegraphics[width=\linewidth]{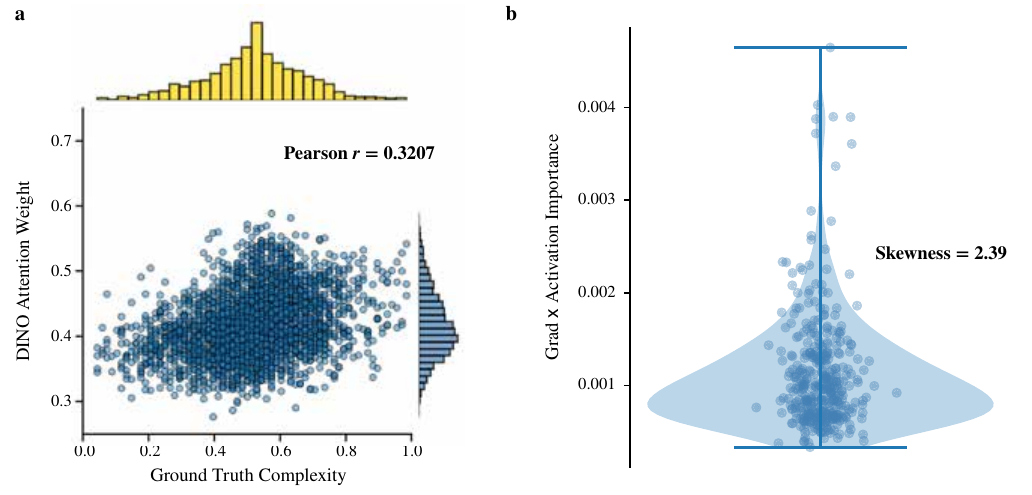}
   \caption{\textbf{Exploration of the DINOv3 [CLS] token in DReX.} 
   (a) Pearson correlation between the attention weight, $w_d$, placed on the DINOv3 [CLS] token and the ground truth complexity score for all images in the IC9600 test set. Marginal histograms are included for each variable. 
   (b) Dimension-level importance of the DINOv3 [CLS] token. The distribution of importance scores is positively skewed ($p < 0.001$).}
  \label{fig:fig3}
\end{figure*}

To identify which components of the pretrained embeddings contribute most to DReX's performance, we conducted feature-level ablation experiments on both branches (Figure~\ref{fig:fig2}c,d). 

For the DINO branch, we ablated each of the 384 dimensions individually by zeroing the $j$-th element of the [CLS] token:
\begin{equation}
\mathbf{f}_{\text{DINO}}^{(-j)} = [f_1, \ldots, f_{j-1}, 0, f_{j+1}, \ldots, f_{384}]
\end{equation}
and measured the resulting change in test set Pearson correlation: $\Delta r_j = r^{(-j)} - r$.

For the ResNet branch, we concatenated the globally pooled features from all four residual blocks to form $\mathbf{f}_{\text{ResNet}} = [\mathbf{f}_1; \mathbf{f}_2; \mathbf{f}_3; \mathbf{f}_4] \in \mathbb{R}^{3840}$, where $\mathbf{f}_i \in \mathbb{R}^{d_i}$ with $d_i \in \{256, 512, 1024, 2048\}$. We ablated each block individually by zeroing all dimensions from that block.

Statistical significance was assessed via paired permutation testing with false discovery rate (FDR) correction at $\alpha = 0.01$. Results reveal that only a small subset of DINO dimensions and mid-to-high level ResNet features (blocks 2--4) significantly impact performance, while block 1 shows minimal effect. This pattern indicates that DReX's fusion mechanism selectively exploits the most informative dimensions from each backbone, with a subset of semantic features from DINO and mid-level structural features from ResNet being most critical for visual complexity prediction.

\subsection{DINOv3 [CLS] Token Exploration}
A novel aspect of DReX is its use of the DINOv3 [CLS] token. While prior methods predominantly rely on convolutional backbones, the 384-dimensional [CLS] token provides a compact global image representation that captures some complementary information about visual complexity. We analyze two properties of the trained fusion module to understand how DReX leverages this representation: (1) how the fusion weight assigned to the DINOv3 branch varies with ground-truth complexity, and (2) how importance is distributed across the 384 dimensions of the [CLS] embedding.

We observe a moderate positive correlation ($r = 0.3207$, Figure~\ref{fig:fig3}a) between the learned DINOv3 attention weight $w_d$ and ground-truth complexity scores. The model adaptively upweights the [CLS] token for more complex images, suggesting that global DINO features provide complementary cues beyond the ResNet-only signal.

To examine dimension-level importance, we compute gradient-activation importance scores:
\begin{equation}
I_j = \frac{1}{N} \sum_{i=1}^{N} \left| \frac{\partial \hat{c}^{(i)}}{\partial f_{\text{DINO}, j}^{(i)}} \cdot f_{\text{DINO}, j}^{(i)} \right|
\end{equation}
where $\hat{c}^{(i)}$ is the predicted complexity score for image $i$, and $f_{\text{DINO}, j}^{(i)}$ is the $j$-th dimension of the [CLS] token. This measures how much each dimension influences the model's predictions. The distribution is positively skewed (Figure~\ref{fig:fig3}b), indicating that DReX concentrates importance on a subset of dimensions. A small number of high-importance dimensions capture the most discriminative complexity-related information, while the majority contribute more modestly.

Together, these analyses demonstrate that the DINOv3 [CLS] token provides a complexity-sensitive signal that complements convolutional features. While not dominant on its own, it is selectively leveraged by DReX when transformer-derived global representations add discriminative value.

\section{Conclusion}
\label{sec:conclusion}

We introduced DReX, a vision-only architecture for predicting human-perceived image complexity that achieves state-of-the-art performance while maintaining parameter efficiency. By fusing self-supervised DINOv3 representations with hierarchical ResNet-50 features through a lightweight attention mechanism, DReX demonstrates that language supervision is not necessary for human-aligned complexity prediction. Our model achieves a Pearson correlation of $r = 0.9581$ on the IC9600 benchmark with approximately $\sim$21.5$\times$ fewer trainable parameters than comparable methods, while simultaneously exhibiting superior generalization across SAVOIAS and PASCAL VOC benchmarks.

The success of DReX rests on several key design principles. By keeping both visual backbones frozen and training only a lightweight fusion module, we demonstrate that the quality of pretrained representations matters more than extensive task-specific fine-tuning. This approach not only improves training efficiency (see Supplementary 3) but also enhances generalization by preventing overfitting to dataset-specific artifacts. The simple, attention-based fusion mechanism adaptively combines complementary visual cues: ResNet's multi-scale hierarchical features capture fine-grained spatial details and texture patterns, while DINOv3's [CLS] token provides global semantic context.

Through systematic ablation studies, we established that both backbone branches contribute meaningfully to DReX's performance. Removing the ResNet branch induces substantial performance degradation ($\Delta r = -0.1192$) while ablating the DINOv3 branch produces a modest but statistically meaningful drop ($\Delta r = -0.0273$), revealing that self-supervised transformer representations provide subtle but complementary information. Our analysis of learned fusion weights shows that the model adaptively upweights DINOv3 features for more complex images ($r = 0.3207$ between attention weight and complexity), indicating context-dependent integration of the two feature streams.

These findings demonstrate that pure vision models can achieve state-of-the-art performance on visual complexity prediction when pretrained representations are appropriately combined, suggesting that language supervision may be unnecessary for tasks grounded primarily in visual perception. More broadly, this lightweight fusion approach---combining frozen foundation models through minimal learned parameters---offers a template for parameter-efficient modeling across diverse perceptual domains.

Beyond visual complexity prediction, DReX's architectural principles may generalize to other perceptual tasks requiring alignment with human judgments, such as image quality assessment and scene understanding. This warrants further investigation. The core insight that frozen pretrained backbones can be efficiently combined through lightweight learned fusion offers a template for developing parameter-efficient, highly generalizable models across diverse perceptual domains and datasets.

While DReX achieves strong performance and insights into the contribution of each backbone branch, several directions warrant exploration. First, our feature ablation studies reveal that importance is concentrated in a subset of dimensions within each backbone, suggesting opportunities for further model compression through principled feature selection or dimensionality reduction while maintaining performance. Characterizing what visual properties these high-importance dimensions encode could reveal which low-level and semantic features are most diagnostic of complexity. Second, extending DReX to predict complexity distributions rather than point estimates could better capture inter-rater variability and uncertainty in human judgments. Third, deploying additional interpretability methods and implementing mechanistic analyses could reveal both what makes individual images complex and how DReX's learned representations relate to cognitive theories of visual complexity, including information-theoretic compressibility and Gestalt principles of perceptual organization. Finally, evaluating DReX across other subjective perceptual dimensions would validate whether its architectural principles generalize beyond complexity prediction to a broader class of human-aligned visual assessment tasks.

In summary, DReX demonstrates that vision-only models built from complementary pretrained representations can achieve state-of-the-art performance on visual complexity prediction while remaining simple, efficient, and highly generalizable—opening promising avenues for lightweight, interpretable solutions to subjective visual assessment tasks.

\section{Code Availability}
Code related to this paper, including a DReX implementation and checkpoint, is available at \texttt{https://github.com/jskaza/DReX}.

\section{Acknowledgments}
Research was sponsored by the U.S. Army Research Office and accomplished under cooperative agreement W911NF-19-2-0026 for the Institute for Collaborative Biotechnologies.

{
    \small
    \bibliographystyle{ieeenat_fullname}
    \bibliography{main}
}

\clearpage
\startfullpagesupplement

\section{Performance Across DINOv3 Backbones}
\label{sec:s1}

\begin{table}[h!]
\caption*{We trained DReX using several DINOv3 backbones while holding all other factors constant, including architecture, initialization seeding, and training procedure. Although larger backbones may benefit from additional training epochs or modified hyperparameters, in our experiments none of the larger models surpassed the performance of the ViT-S/16 backbone.}
\centering
\begin{tabularx}{\textwidth}{Xcccc}
\toprule
\textbf{DINO Backbone} & \textbf{Parameters}  & \textbf{Embedding Dimension} & \textbf{Pearson $r$}  & \textbf{Spearman's $\rho$} \\
\midrule
ViT-S/16   & 21M & 384  & 0.9581  & 0.9542 \\
ViT-S+/16  & 29M & 384  & 0.9576  & 0.9537 \\
ViT-B/16   & 86M & 768  & 0.9553  & 0.9519 \\
ViT-L/16   & 300M & 1024 & 0.9556  & 0.9513 \\
ViT-H+/16  & 840M & 1280 & 0.9514  & 0.9482 \\
\bottomrule
\end{tabularx}
\end{table}

\newpage
\section{Model Performance Across Seeds}
\label{sec:s2}
\begin{figure*}[!h]
  \centering
   \includegraphics[width=5.5in]{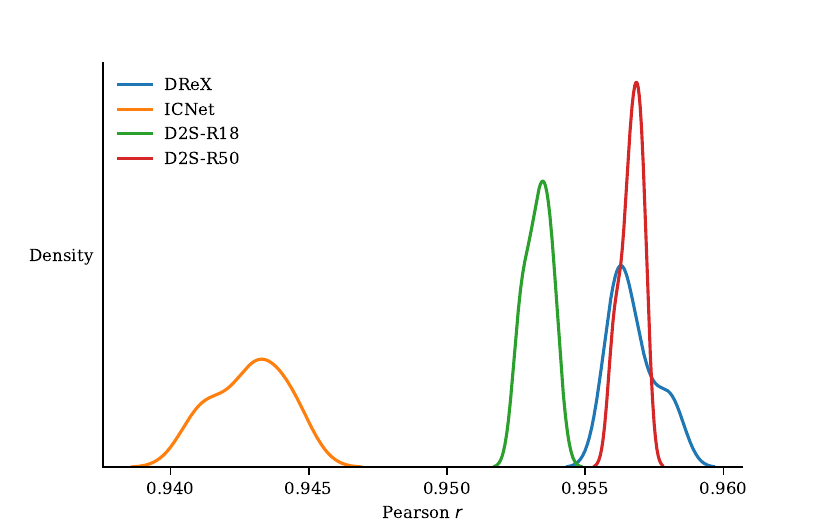}
   \includegraphics[width=5.5in]{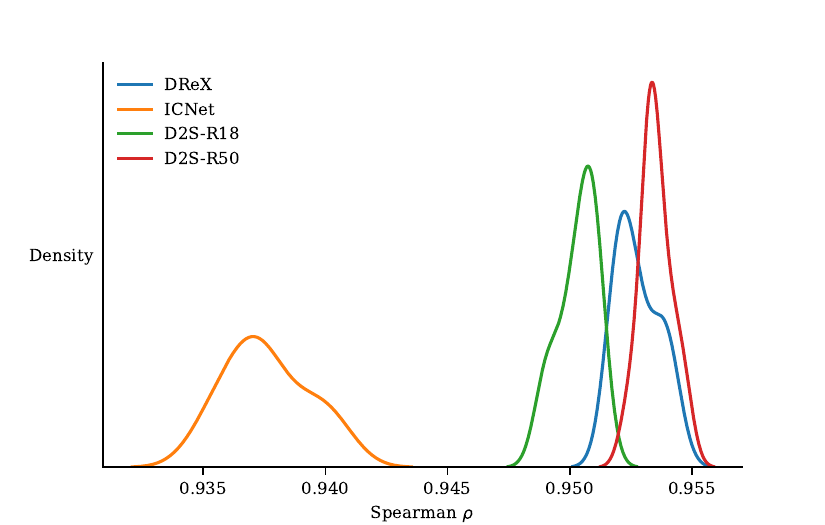}
   \caption*{Kernel density estimation of Pearson correlation ($r$) and Spearman's rank correlation ($\rho$) across 10 different random seeds for the models DReX, ICNet, D2S-R50, and D2S-R18.}
\end{figure*}

\newpage
\section{Compute Savings During Training}
\label{sec:s3}
\begin{figure*}[!h]
  \centering
   \includegraphics[width=6.5in]{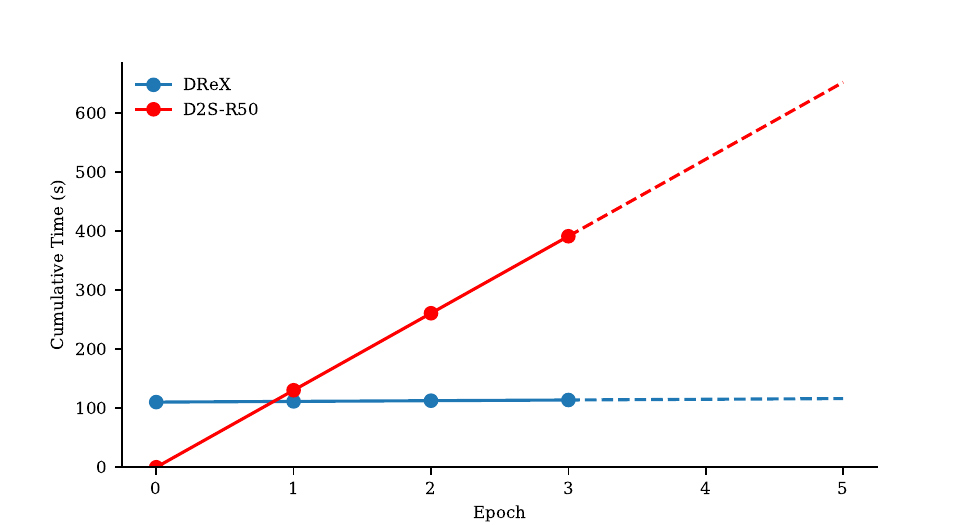}
   \caption*{Comparison of training time between DReX and D2S-R50. Using DReX, we precompute ResNet and DINOv3 features once and reuse them across all training epochs, which significantly reduces the time per epoch. The initial precomputation takes approximately 110~s for the IC9600 training set, after which each epoch takes only $\sim$1.2~s. In contrast, each epoch of D2S-R50 requires $\sim$130.5~s. These estimates were obtained on a workstation featuring an AMD Ryzen Threadripper 7970X 32-core processor and an NVIDIA RTX Pro 6000 Blackwell Workstation GPU.}
\end{figure*}

%

\end{document}